\documentclass[twoside,11pt]{article}

\usepackage{jmlr2e}

\ShortHeadings{Pure Exploration in Infinite Bandit Models}{Aziz et al.}
\firstpageno{1}

\usepackage[utf8]{inputenc} 
\usepackage{hyperref}       
\usepackage{url}            
\usepackage{booktabs}       
\usepackage{amsfonts}       
\usepackage{nicefrac}       
\usepackage{microtype}      

\usepackage{algorithmic}
\usepackage{algorithm2e}

\usepackage{amsfonts}
\usepackage{amsmath}
\usepackage{amssymb}
\usepackage{dsfont}
\usepackage{enumitem}
\usepackage{filecontents}
\usepackage{todonotes}
\usepackage{graphicx}
\usepackage{subcaption}
\usepackage{wrapfig}

\newtheorem{assumption}{Assumption}
\newcommand*{\qed}{\hfill\ensuremath{\square}}%

\newcommand{\diff}{\mathop{}\!\mathrm{d}}

\DeclareMathOperator*{\argmax}{argmax}
\newcommand{\PrNone}{\mathbb{P}}
\renewcommand{\Pr}[1]{\PrNone\left( #1 \right)}
\newcommand{\ENone}{\mathds{E}}
\newcommand{\E}[1]{\ENone\left[ #1 \right]}
\newcommand{\KL}[2]{\operatorname{KL}\left( #1 \| #2 \right)}
\newcommand{\R}{\mathbb{R}}
\newcommand{\N}{\mathbb{N}}
\newcommand{\1}[1]{\mathds{1}_{\left( #1 \right)}}
\newcommand{\Set}[1]{\mathchoice%
{\left\{ #1 \right\}}{\{ #1 \}}{\{ #1 \}}{\{ #1 \}}}

\newcommand{\BernKL}[2]{\operatorname{kl}\left( #1, #2 \right)} 

\newcommand{\historyRV}{H}
\newcommand{\armRV}{W}
\newcommand{\rewardRV}{Z}

\newcommand{\Arms}{\mathcal{W}}
\newcommand{\arm}{w}
\newcommand{\ArmEvents}{\mathcal{F}_\Arms}
\newcommand{\ArmMeasure}{M}
\newcommand{\ArmMean}{\mu}
\newcommand{\ArmMeanAlt}{\lambda}
\newcommand{\ArmGrp}[1]{\mathcal{S}^{#1}}
\newcommand{\ArmGrpEmp}[1]{\mathcal{S}_{#1}}
\newcommand{\ArmKLNone}{\operatorname{d}}
\newcommand{\ArmKL}[2]{\ArmKLNone\hspace{-.1em}\left( #1, #2 \right)}
\newcommand{\ArmCI}[2]{\operatorname{d^*}\hspace{-.2em}\left( #1, #2 \right)}

\newcommand{\ExpRewards}{\Theta}
\newcommand{\expReward}{\theta}
\newcommand{\ExpRewardEvents}{\mathcal{F}_\ExpRewards}
\newcommand{\ExpRewardMeasure}{M_\ExpRewards}
\newcommand{\ExpRewardDensity}{g}
\newcommand{\ExpRewardCDF}{G}
\newcommand{\ExpRewardInvCDF}{G^{-1}}

\newcommand{\GTargetEps}{\mathcal{G}_{\ArmMeasure,\ArmMean}^{\alpha,\epsilon}}
\newcommand{\GTargetZeroEps}{\mathcal{G}_{\ArmMeasure,\ArmMean}^{\alpha,0}}
\newcommand{\GTargetNoEps}{\mathcal{G}_{\ArmMeasure,\ArmMean}^{\alpha}}
\newcommand{\GTargetNoEpsAlt}{\mathcal{G}_{\ArmMeasure,\ArmMeanAlt^i}^{\alpha}}
\newcommand{\GTargetNoEpsAltOne}{\mathcal{G}_{\ArmMeasure,\ArmMeanAlt^1}^{\alpha}}

\newcommand{\PrTrueModelNone}{\PrNone_{\ArmMean}^\ArmMeasure}
\newcommand{\PrTrueModel}[1]{\PrNone_{\ArmMean}^\ArmMeasure\left( #1 \right)}
\newcommand{\PrAltModel}[1]{\PrNone_{\ArmMeanAlt}^\ArmMeasure\left( #1 \right)}
\newcommand{\PrAltModelI}[1]{\PrNone_{\ArmMeanAlt^i}^\ArmMeasure\left( #1 \right)}
\newcommand{\PrAltModelOne}[1]{\PrNone_{\ArmMeanAlt^1}^\ArmMeasure\left( #1 \right)}

\newcommand{\ETrueModelNone}{\ENone_{\ArmMean}^\ArmMeasure}
\newcommand{\ETrueModel}[1]{\ENone_{\ArmMean}^\ArmMeasure\left[ #1 \right]}
\newcommand{\EAltModel}[1]{\ENone_{\ArmMeanAlt}^\ArmMeasure\left[ #1 \right]}

\newcommand{\ArmInterval}{\mathcal{I}_{a}(t)}

\newcommand{\EpsCheapArms}{\mathcal{A}_{\epsilon}}
\newcommand{\cE}{\mathcal{E}}
\newcommand{\cF}{\mathcal{F}}
\newcommand{\cR}{\mathcal{R}}

\usepackage{bm}

\catcode`@=11

% ...

\def\imparray{\stepcounter{equation}\let\@currentlabel=\theequation
\global\@eqnswtrue
\global\@eqcnt\z@\tabskip\@centering\let\\=\@eqncr
$$\halign to \displaywidth\bgroup\llap{${##}$\hskip 4\arraycolsep}\tabskip\z@&
  \@eqnsel\hskip\@centering
  $\displaystyle\tabskip\z@{##}$&\global\@eqcnt\@ne 
  \hskip 2\arraycolsep \hfil${##}$\hfil
  &\global\@eqcnt\tw@ \hskip 2\arraycolsep $\displaystyle\tabskip\z@{##}$\hfil 
   \tabskip\@centering&\llap{##}\tabskip\z@\cr}

\def\endimparray{\@@eqncr\egroup
      \global\advance\c@equation\m@ne$$\global\@ignoretrue}

\@namedef{imparray*}{\def\@eqncr{\nonumber\@seqncr}\imparray}
\@namedef{endimparray*}{\nonumber\endimparray}

\catcode`@=12

\title{Pure Exploration in Infinitely-Armed Bandit Models with Fixed-Confidence}
\author{\name Maryam Aziz \email azizm@ccs.neu.edu \\
  \addr Northeastern University\\
  Boston, MA
  \AND
  \name Jesse Anderton \email jesse@ccs.neu.edu \\
  \addr Northeastern University\\
  Boston, MA
  \AND
  \name Emilie Kaufmann \email emilie.kaufmann@univ-lille1.fr \\
\addr CNRS \& CRIStAL, UMR 9189, Universit\'e de Lille \\
Inria Lille, SequeL team
  \AND
  \name Javed Aslam \email jaa@ccs.neu.edu \\
  \addr Northeastern University\\
  Boston, MA
}
\editor{}

\begin{document}

\maketitle

\begin{abstract}%

We consider the problem of near-optimal arm identification in the fixed confidence setting of the infinitely armed bandit problem
when nothing is known about the arm reservoir distribution.
We 
(1)~introduce a PAC-like framework
within which to derive and cast results;
(2)~derive a sample complexity lower bound for near-optimal arm
identification;
(3)~propose an algorithm that identifies a nearly-optimal arm with high probability
and derive an upper bound on its sample complexity which is within a log factor of our
lower bound; and
(4)~discuss whether our $\log^2 \frac{1}{\delta}$ dependence is inescapable for ``two-phase'' (select arms first, identify the best later) algorithms in the infinite setting.
This work permits the application of bandit models to a broader class of
problems where fewer assumptions hold.
\end{abstract}



%
%
\section{Introduction}\label{sec-intro}

We present an extension of the \emph{stochastic multi-armed
bandit} (MAB) model, which is applied to
many problems in computer science and beyond.  In a bandit model,
an agent
is confronted with a set of arms
that are unknown probability distributions.
At each round $t$, the agent chooses an arm to play,
based on past observation, after which a reward
drawn from the arm's distribution is observed. This sequential
sampling strategy (``bandit algorithm'') is adjusted to
optimize some utility measure.
Two measures are typical: cumulative regret
minimization and pure exploration.
For regret minimization, one attempts
 minimize \emph{regret}, the difference between
the expected cumulative rewards of an
optimal strategy and the employed strategy.
In the pure-exploration framework, one seeks
the arm with largest mean irrespective of the observed rewards.
Two dual settings have been studied: the \emph{fixed-budget}
setting, wherein one
can use
only a given number of arm-pulls,
and the \emph{fixed-confidence} setting, wherein one
attempts to achieve a utility target with minimal arm-pulls.


While the literature mainly considers bandit models with a known,
\emph{finite} number of arms, for many applications the number of arms may
be very large and even infinite.
In these cases, one can often settle for an arm which is
``near'' the best in some sense, as such an arm can be identified at
significantly less cost.
One such application is machine learning: given a large pool of
possible classifiers (arms),
one wants to find the one with minimal risk (mean reward) by sequentially
choosing a classifier, training it
and measuring its empirical test
error (reward).
In text, image, and video classification, one often encounters effectively
infinite sets of classifiers which are prohibitively expensive to
assess individually.
Addressing such cases with bandit models
is particularly useful when used within ensemble algorithms such as
AdaBoost
\citep{Freund:1996:ENB:3091696.3091715}, and some variations on this idea have already
been explored
\citep{icml2013_appel13, busafekete:in2p3-00614564,
Dubout:2014:ASL:2627435.2638580, Escudero:2001:ULW:2387364.2387381}, though the
task of efficiently identifying a near-optimal classifier is at present
unsolved.
We here approach such problems from a theoretical standpoint.


Two distinct lines of work address a
potentially \emph{infinite set of arms}. Let $\Arms$ be a (potentially
uncountable) set of arms and assume that there exists $\ArmMean : \Arms \rightarrow
\R$, a mean-reward mapping such that when some arm $\arm$ is selected, one observes
an independent draw of a random variable with mean $\ArmMean(\arm)$.
One line of research
\citep{Kleinberg08Infinite,Bubeck11Xarmed,grill2015black-box}
assumes that $\Arms$ is some
metric space, and that $\ArmMean$ has some regularity property with respect to the
metric (for example it is locally-Lipschitz). Both regret minimization and
fixed-budget pure-exploration problems have been studied in this setting.
Another line of research, starting with the work
of \cite{berry1997} assumes no particular structure on $\Arms$ and no regularity
for $\ArmMean$. Rather, there is some \emph{reservoir distribution} on the arms'
means (the set $\ArmMean(\Arms)$ with our notation) such that at each round the
learner can decide to query a new arm, whose mean is drawn from the reservoir,
and sample it, or to sample an arm that was previously queried. While regret
minimization was studied by several authors~\citep{NIPS2008_3452,
NIPS2013_5109,10.1007/978-3-662-44848-9_20}, the recent work
of~\cite{DBLP:journals/corr/CarpentierV15} is the first to study the
pure-exploration problem in the fixed-budget setting.


We present a novel theoretical framework for the
fixed-confidence pure-exploration problem in an infinite bandit model with a
reservoir distribution.
The reservoir setting seems well-suited for machine learning, since it is not clear whether the
test error of a parametric classifier is smooth with respect to its parameters.
Typically, an assumption is made on the form of the tail of the reservoir which allows
estimation of the probability that an independently-drawn arm will be
``good;'' that is, close to the best possible arm.
However, for problems 
such as that mentioned above
 such an assumption does not seem warranted.
Instead, we employ a parameter, $\alpha$, indicating the probability of
independently drawing a ``good'' arm.
When a tail assumption can be made, $\alpha$ can be computed from this
assumption. Otherwise, it can be chosen based on the user's needs.
Note that the problem of identifying a ``top-$\alpha$'' arm in the infinite
case corresponds to the finite case problem of
finding one of the top $m$ arms from a set of $n$ arms, for $n>m$,
with $\alpha = m/n$.
The first of two PAC-like frameworks we introduce, the $(\alpha,\delta)$
framework,
aims to identify an arm in the top-$\alpha$ tail of the reservoir with 
probability at least $1-\delta$, using as few samples as
possible.

We now motivate our second framework.
When no assumptions can be made on the reservoir, one may encounter
reservoirs with large probability masses close to the boundary of
the
top-$\alpha$ tail.
Indeed, the distribution of weighted classifier accuracies in later rounds of 
AdaBoost has this property, as the weights are chosen to drive all 
classifiers toward random performance.
This is a problem for any framework defined purely in terms of $\alpha$, because
such masses make us likely to observe arms which are
not in the top-$\alpha$ tail but which are hard to distinguish from
top-$\alpha$ arms.
However, in practice their similarity to top-$\alpha$ arms makes them reasonable arms to
select.
For this reason we add an $\epsilon$ relaxation, which
limits the effort spent on
arms near the top-$\alpha$ tail
while adding directly to the simple regret a user may observe.
Formally, our $(\alpha,\epsilon, \delta)$ framework seeks
an arm within $\epsilon$ of the top-$\alpha$ fraction of the arms with
probability at least $1-\delta$,
using as few samples from the arms as possible.

Although $\alpha$ and $\epsilon$ both serve to relax
the goal of finding an arm with maximum mean,
they have distinct purposes and are both useful.
One might wonder,
if the inverse CDF $\ExpRewardInvCDF$ for the arm reservoir was available
(at least at the tail), why one would not simply compute
$\epsilon'=\epsilon+\ExpRewardInvCDF(1-\alpha)$ and use the
established $(\epsilon,\delta)$ framework.
Indeed, $\alpha$ is important precisely when 
the form of the reservoir tail is unknown.
The user of an algorithm will wish to limit the effort spent in finding an optimal arm,
and with no assumptions on the reservoir $\epsilon$ alone is insufficient
to limit an algorithm's sample complexity.
Just as there might be large probability close to the $\alpha$ boundary,
it may be that there is virtually no probability within $\epsilon$ of
the top arm.
The user applies
$\alpha$ to (effectively) specify how hard to work to
estimate the reservoir tail,
and $\epsilon$ to specify how hard to work to
differentiate between individual arms.

Our approach
differs from the typical reservoir setting in that it does not require any
regularity assumption on the tail of the reservoir distribution,
although it can take advantage of one when available.
Within this framework, we prove a lower bound on the expected number
of arm pulls necessary to achieve $(\alpha,\delta)$
or $(\alpha,\epsilon,\delta)$
performance by generalizing the information-theoretic tools introduced
by~\cite{JMLR15}
in the finite MAB setting.  We also study
a simple algorithmic solution to the problem based on 
the \texttt{KL-LUCB} algorithm of \cite{COLT13}, an
algorithm for $\epsilon$-best arm identification in bandit models with a finite
number of arms, and we compare its performance to our derived
lower bound theoretically.
Our algorithm is an $(\alpha, \epsilon, \delta)$ algorithm,
but we show how to achieve $(\alpha,\delta)$ performance
when assumptions can be made on the tail of the reservoir.

We introduce
the $(\alpha,\delta)$ and $(\alpha, \epsilon, \delta)$ frameworks and relate
them to existing literature in Section~\ref{sec-prelim}.
Section~\ref{sec-fc-lower} proves our sample complexity lower bounds.
In  Section~\ref{sec-fc-upper}, we present and analyze the \texttt{$(\alpha,\epsilon)$-KL-LUCB} algorithm for one-dimensional exponential family reward distributions.
A comparison between our upper and lower bounds can be found in Section~\ref{sec-compare-lb-ub}.
We defer most proofs to the appendix, along with some
numerical experiments.

%
%
\section{Pure Exploration with Fixed Confidence}\label{sec-prelim}

Here we formalize our frameworks and connect them to the existing literature.


\subsection{Setup, Assumptions, and Notation}\label{setup}

Let $(\Arms, \ArmEvents, \ArmMeasure)$ be a probability space over arms
with measure $\ArmMeasure$,
where each arm $\arm \in \Arms$ is some abstract object (e.g. a classifier),
and let $(\ExpRewards, \ExpRewardEvents)$ be a measurable space over
expected rewards, where $\ExpRewards \subseteq \R$ is a
continuous interval
and $\ExpRewardEvents$ is the Borel $\sigma$-algebra over $\ExpRewards$
(i.e. the smallest $\sigma$-algebra containing all sub-intervals of $\Theta$).
Also let
$P_{\ExpRewards} = \Set{ p_{\expReward}, \expReward \in \ExpRewards }$
be a parametric set of probability distributions such that each
distribution is continuously parameterized by its mean.
To ease the notation, we shall assume
$\ENone_{X \sim p_{\expReward}}[X] = \expReward$.
One can think of $P_{\ExpRewards}$ as a one-parameter exponential family (e.g. the
family of Bernoulli, Gaussian with fixed and known variance, Poisson or Exponential distributions with means
in some interval or other subset of $\R$),
however we do not limit ourselves to such well-behaved reward distributions.
We defer our further assumptions on $P_\Theta$ to Section~\ref{sec-fc-lower-stmt}.
We will denote by $f_{\expReward}$ the density of the element in $P_{\ExpRewards}$
with mean $\expReward$.

An \emph{infinite bandit model} is characterized by a probability measure
$\ArmMeasure$ over $(\Arms, \ArmEvents)$ together with a
measurable mapping $\ArmMean : \Arms \rightarrow \ExpRewards$
assigning a mean $\expReward$
(and therefore a reward distribution $p_\expReward$) to each arm.
The role of the measure $\ArmMeasure$ is to define the top-$\alpha$ fraction of arms, as we will show in Eq.~\ref{eq-good-arms};
it can be used by the algorithm to sample arms.
At each time step $t$, a user
selects an arm $\armRV_t \in \Arms$, based on past observation. He
can either query a new arm in $\Arms$
(which may be sampled $\armRV_t \sim \ArmMeasure$, or selected adaptively) or select an arm that
has been queried
in previous rounds. In any case, when arm $\armRV_t$ is drawn, an independent sample
$\rewardRV_t \sim p_{\ArmMean(\armRV_t)}$ is observed. 

For example, when boosting decision stumps
(binary classifiers which test a single feature against a threshold),
the set of arms $\Arms$ consists of all possible decision
stumps for the corpus, and
the expected reward for each arm is its expected accuracy over the sample space 
of all possible classification examples.
An algorithm may choose to draw the arms at random according to the
probability measure $\ArmMeasure$;
this is commonly done by, in effect, placing uniform probability mass over the
thresholds placed halfway between the distinct values seen in the training data
and placing zero mass over the remaining thresholds.
We are particularly interested in the case when the number of arms in the
support for $\ArmMeasure$ is so large as to be effectively infinite, at least with
respect to the available computational resources.

We denote by $\PrTrueModelNone$ and $\ETrueModelNone$ the probability and 
expectation under an infinite bandit model with arm probability measure
$\ArmMeasure$ and mean function $\ArmMean$. The
history of the bandit game up to time $t$ is
$\historyRV_t = ((\armRV_1,\rewardRV_1),\dots,(\armRV_t,\rewardRV_t)).$
By our assumption, the arm selected at round $t$ only depends on
$\historyRV_{t-1}$
and $U_t$, which is uniform on $[0,1]$ and independent of
$\historyRV_{t-1}$ (used to sample from $\ArmMeasure$ if needed).
In particular, the
conditional density of $\armRV_t$ given $\historyRV_{t-1}$,
denoted by  $\PrTrueModel{ \armRV_t | \historyRV_{t-1} }$,
is independent of the mean mapping $\ArmMean$. Note that this
property is
satisfied as well if, when querying a new arm, $\armRV_t$ can be chosen
arbitrarily 
in $\Arms$ (depending on $\historyRV_{t-1}$), and not necessarily at random from
$\ArmMeasure$. Under these
assumptions, one can compute the likelihood of $\historyRV_T$:
\vspace{-10pt}
\begin{align}
\ell\left(\historyRV_T ; \ArmMean,\ArmMeasure\right) = \prod_{t=1}^T f_{\ArmMean(\armRV_t)}(\rewardRV_t)
\PrTrueModel{ \armRV_t | \historyRV_{t-1} }.\label{Likelihood}
\end{align}
Note that the arms $\armRV_t$ are not latent objects: they are assumed to be
observed, but not their means $\ArmMean(\armRV_t)$.
For instance, in our text classification example we know the classifier we are
testing but not its true classification accuracy.
Treating arms as observed in this way simplifies the likelihood by making the
choice of new arms to query independent of their mean mappings.
This is key to our approach to dealing with reservoirs about which nothing
is known; we can avoid integrating over such reservoirs and so do not require
the reservoir to be smooth.
For details, see Appendix~\ref{proof-CD}.

\subsection{Objective and Generic Algorithm}\label{generic}

\paragraph{Reservoir distribution.}
The probability space over arms $(\Arms, \ArmEvents, \ArmMeasure)$
and the mapping $\ArmMean$ is used to
form a pushforward measure over expected rewards
$
\ExpRewardMeasure(\cE) := (\ArmMean_*(\ArmMeasure))(\cE)
	= \ArmMeasure(\ArmMean^{-1}(\cE)), \text{ for } \cE \in \ExpRewardEvents,
$
inducing the probability space
$(\ExpRewards, \ExpRewardEvents, \ExpRewardMeasure)$
over expected rewards.
We define our \emph{reservoir distribution} CDF
$
	\ExpRewardCDF(\tau) =
		\ExpRewardMeasure( \Set{\theta \le \tau} )
$
whose density $\ExpRewardDensity$ is its Radon-Nikodym derivative
with respect to $\ArmMeasure$.
For convenience, we also define the ``inverse'' CDF
$
\ExpRewardInvCDF(p) := \inf \Set{ \expReward : G(\expReward) \ge p }.
$
%
We assume that $G$ has bounded support and let
$\ArmMean^*$ be the largest possible mean under the reservoir distribution,
$
\ArmMean^* := \ExpRewardInvCDF(1)
	= \inf \Set{\expReward : G(\expReward) = 1 }.\label{eq-mustar}
$

In the general setup introduced above, the reservoir may or may not be 
useful to query new arms, but it is needed to define the notion of top-$\alpha$
fraction.

\paragraph{Finding an arm in the top-$\alpha$ fraction.}
In our setting, for some fixed $\alpha \in ]0,1[$ and some $\epsilon \geq 0$, the goal is to identify an
arm that belongs to the set
\begin{align}\label{eq-good-arms}
\GTargetEps := \Set{
\arm \in \Arms :
\ArmMean(\arm) \ge \ExpRewardInvCDF(1 - \alpha) - \epsilon}
\end{align}
of arms whose expected mean rewards is high, in the sense that their mean is
within $\epsilon$ of
the quantile of order $1-\alpha$ of the reservoir distribution.
For notational convenience, when we set $\epsilon$ to zero we write 
$\GTargetNoEps := \GTargetZeroEps$.

\begin{wrapfigure}{r}{0.4\textwidth}
\vspace{-15pt}
\begin{minipage}{0.4\textwidth}
\begin{algorithm}[H]
\caption{Generic algorithm.}
\label{algo:Generic}
\begin{algorithmic}
\REQUIRE{Arm set $\Arms$, target $\alpha,\epsilon,\delta$}
\ENSURE{Some arm $\hat{s}$}
\FOR{$t \gets 1,2,\dots$}
 \STATE(choose one of:){
 \begin{enumerate}
 \item Pull arm: Choose $\armRV_t \sim \PrTrueModel{\armRV_t |
H_{t-1}}$
     and observe reward $\rewardRV_t \sim p_{\ArmMean(\armRV_t)}$
 \item Stop: Choose $\hat{s} \gets \armRV_s$ for some $s < t$, \\
	{\textbf{return} $\hat{s}$} 
\end{enumerate}
 }
\ENDFOR
\end{algorithmic}
\end{algorithm}
\end{minipage}
\vspace{-15pt}
\end{wrapfigure}

\paragraph{Generic algorithm}  An algorithm is made of a
\emph{sampling rule} $(\armRV_t)$, a \emph{stopping rule} $\tau$ (with
respect to the filtration generated by $\historyRV_t$) and a \emph{recommendation rule}
$\hat{s}_\tau$ that selects one of the queried arms as a candidate arm from
$\GTargetEps$. This is summarized in
Algorithm~\ref{algo:Generic}.

Fix $\delta \in ]0,1[$.
An algorithm that returns an arm from $\GTargetEps$
with probability at least $1-\delta$ is said to be
$(\alpha,\epsilon,\delta)$-correct.
Moreover, an $(\alpha,\epsilon,\delta)$-correct algorithm must perform
well on
all possible infinite bandit models:
$
\forall (\ArmMeasure, \ArmMean), \ \
\PrTrueModel{ \hat{s}_\tau \in \GTargetEps }
	\geq 1 - \delta.
$
Our
goal is to build an $(\alpha,\epsilon,\delta)$-correct algorithm that uses as
few
samples as possible, i.e. for which $\ETrueModel{\tau}$ is small.
We similarly define the notion of $(\alpha,\delta)$-correctness
when
$\epsilon=0$.

\paragraph{$(\alpha, \epsilon, \delta)$-correctness}
When little is known about the reservoir distribution
(e.g. it might not even be smooth), an $\epsilon$-relaxed algorithm is 
appropriate.
The choice of $\epsilon$ represents a tradeoff between simple regret
(defined shortly)
and the maximum budget used to differentiate between arms.
We provide our lower bound in both $\epsilon$-relaxed and unrelaxed forms.
Our algorithm requires an $\epsilon$ parameter, but we show how this parameter
can be chosen under regularity assumptions on the tail of the reservoir to
provide an $(\alpha,\delta)$-correct algorithm.

\paragraph{Simple regret guarantees.} In the infinite bandit literature,
performance is typically measured in
terms of
\emph{simple regret}: $r_\tau = \ArmMean^* - \ArmMean(\hat{s}_\tau)$.
If the tail of the reservoir distribution is bounded, one
can obtain simple regret upper bounds for an
algorithm in our framework.

A classic assumption (see, e.g. \cite{DBLP:journals/corr/CarpentierV15}) is
that there
exists $\beta >0$ and two constants $E,E'$ such that
\vspace{-10pt}
\begin{align}
\forall \rho >0,
	E\rho^\beta \leq \ExpRewardMeasure(\Set{\expReward
\geq \ArmMean^* - \rho})
	\leq E'\rho^\beta.\label{ass:smoothness}
\end{align}
With $C = E^{-1/\beta}$ and $C'=(E')^{-1/\beta}$, this translates into 
$
\forall \alpha >0, C\alpha^{1/\beta}
	\leq \ArmMean^* - \ExpRewardInvCDF(1-\alpha)
	\leq C' \alpha^{1/\beta},
$
and a $(\alpha,\delta)$-correct algorithm has its simple regret upper bounded as
\vspace{-10pt}
\begin{align}
\Pr{ r_\tau \leq C'\alpha^{1/\beta} } \geq 1-\delta.
\end{align}
Similarly, a $(\alpha,\epsilon,\delta)$-correct algorithm has its simple regret bounded
as
\vspace{-10pt}
\begin{align}
\Pr{ r_\tau \leq C'\alpha^{1/\beta} + \epsilon }
	\geq 1-\delta.
\end{align}
If
$\beta$ is known, $\alpha$ and $\epsilon$
can be chosen to guarantee a simple regret below an arbitrary bound.

\subsection{Related Work}\label{sec-related}

Bandit models were introduced by
\cite{10.2307/2332286}.
There has been recent interest in
pure-exploration problems
\citep{EvenDaral06,Bubeck10BestArm}; for which good algorithms
are expected to differ from those for the classic
regret minimization objective
\citep{Bubeckal11,ESAIM17}.

For a finite number of arms with means $\ArmMean_1,\dots,\ArmMean_K$, the fixed-confidence
best arm identification problem was introduced by \cite{EvenDaral06}. The goal
is to
select an arm $\hat{a} \in \{1,\dots,K\}$ satisfying
$\Pr{ \ArmMean_{\hat{a}} \geq \ArmMean^* - \epsilon } \geq 1-\delta$, where
$\ArmMean^* = \max_a \ArmMean_a$. Such an algorithm is called $(\epsilon,\delta)$-PAC.
In our setting, assuming a uniform reservoir distribution over
$\{1,\dots,K\}$ yields an $(\alpha,\delta)$-correct algorithm with $\alpha$
being the fraction of $\epsilon$-good arms. 
Algorithms are either
based on successive eliminations \citep{EvenDaral06,icml2013_karnin13} or on 
confidence intervals
\citep{DBLP:conf/icml/KalyanakrishnanTAS12,NIPS2012_4640}. For exponential family reward
distributions, the \texttt{KL-LUCB}
algorithm of \cite{COLT13} refines the confidence
intervals to obtain better performance compared to its Hoeffding-based
counterpart, and a sample complexity scaling with the Chernoff information
between arms
(an information-theoretic measure related to the Kullback-Leibler
divergence). We build on this algorithm to define \texttt{$(\alpha,\epsilon)$-KL-LUCB} in
Section~\ref{sec-fc-upper}. \emph{Lower bounds} on the sample complexity have
also
been proposed by \cite{Mannor04thesample,JMLR15,GK16}.
 In Section~\ref{sec-fc-lower} we generalize the change of
distribution tools used therein to present a lower bound for pure exploration
in an infinite bandit model.

Regret minimization has been studied
extensively for infinite bandit models
\citep{berry1997,NIPS2008_3452, NIPS2013_5109,
10.1007/978-3-662-44848-9_20}, whereas
\cite{DBLP:journals/corr/CarpentierV15} is the first work dealing with
pure-exploration for general reservoirs. The authors consider the fixed-budget
setting, under the tail
assumption \eqref{ass:smoothness} for the reservoir distribution,
already discussed.

Although the fixed-confidence pure-exploration problem for infinitely armed 
bandits has been rarely addressed
for general reservoir distributions, the \emph{most-biased
coin problem} studied by \cite{DBLP:journals/corr/abs-1202-3639,
Jamieson:2016wd} can be viewed as a particular
instance, with a specific reservoir distribution that is a mixture of
``heavy'' coins of mean $\expReward_1$ and ``light'' coins
of mean $\expReward_0$: $G = (1-\alpha)\delta_{\expReward_0} + \alpha
\delta_{\expReward_1}$, where $\expReward_1 > \expReward_0$
and with $\delta_\theta$ here denoting the Dirac delta function. The goal is to
identify, with probability at least $1-\delta$, an arm with mean
$\expReward_1$. If a lower bound $\alpha_0$ on $\alpha$ is known, this
is equivalent to finding an $(\alpha_0,\delta)$-correct algorithm
by our definition.
We suggest in Section~\ref{sec-compare-lb-ub} that the
sample complexity of any two-phase algorithm (such as ours) might scale like $\log^2 \frac{1}{\delta}$, while Jamieson et al. achieve
a dependence on $\delta$ of $\log \frac{1}{\delta}$ for the special case they address.

Finally, the recent work of \cite{Chaudhuri2017PACIO} studies a framework that is similar to the one introduced in this paper\footnote{Note that we became aware of their work after submitting our paper.}. Their first goal of identifying, in a finite bandit model an arm with mean larger than $\mu_{[m]} - \epsilon$ (with $\mu_{[m]}$ the arm with $m$-th largest mean) is extended to the infinite case, in which the aim is to find an $(\alpha,\epsilon)$-optimal arm. 
The first algorithm proposed for the infinite case applies the Median Elimination algorithm \citep{EvenDaral06} on top of
$\frac{1}{\alpha}\log\frac{2}{\delta}$ 
arms drawn from the reservoir and is proved to have a
$O\left(\frac{1}{\alpha\epsilon^2} \log^2\frac{1}{\delta}\right)$
sample complexity. The dependency in $\log^2\frac{1}{\delta}$ is the same as the one we obtain for \texttt{$(\alpha,\epsilon)$-KL-LUCB}, however our analysis goes beyond the scaling in $\frac{1}{\epsilon^2}$ and reveals a complexity term based on KL-divergence, that can be significantly smaller. 
Another algorithm is presented, without sample complexity guarantees,
that runs \texttt{LUCB} on successive batches of arms drawn from the reservoir in order to avoid memory storage issues.



\section{Lower Bound}\label{sec-fc-lower}

We now provide sample complexity lower bounds for our two frameworks.

\subsection{Sample complexity lower bound}\label{sec-fc-lower-stmt}

Our lower bound scales with the Kullback-Leibler divergence between arm
distributions $p_{\expReward_1}$ and $p_{\expReward_2}$, denoted by
$
\ArmKL{\expReward_1}{\expReward_2} := \KL{p_{\expReward_1}}{p_{\expReward_2}}
	= \ENone_{X \sim p_{\expReward_1}}\left[\log \frac{f_{\expReward_1}(X)}{f_{\expReward_2}(X)}\right].
$

Furthermore, we make the following assumptions on the arm reward distributions,
that are typically satisfied for one-dimensional exponential families.
\begin{assumption}\label{ass:kldiv}
The KL divergence, that is the application
$(\expReward_1,\expReward_2) \mapsto \ArmKL{\expReward_1}{\expReward_2}$
is continuous on $\ExpRewards \times \ExpRewards$,
and $\ExpRewards$ and $\ArmKLNone$ satisfy 
\begin{itemize}[topsep=0pt,itemsep=0pt]
\item $\expReward_1 \ne \expReward_2 \Rightarrow
    0 < \ArmKL{\expReward_1}{\expReward_2} < \infty$
\item $\expReward_1 < \expReward_2 < \expReward_3 \Rightarrow
    \ArmKL{\expReward_1}{\expReward_3} >  
    \ArmKL{\expReward_2}{\expReward_3}$
    and $\ArmKL{\expReward_1}{\expReward_2} <  
    \ArmKL{\expReward_1}{\expReward_3}$
\end{itemize}
\end{assumption}
 
\begin{wrapfigure}{r}{0.35\textwidth}
	\vspace{-20pt}
	\includegraphics[width=0.35\textwidth]{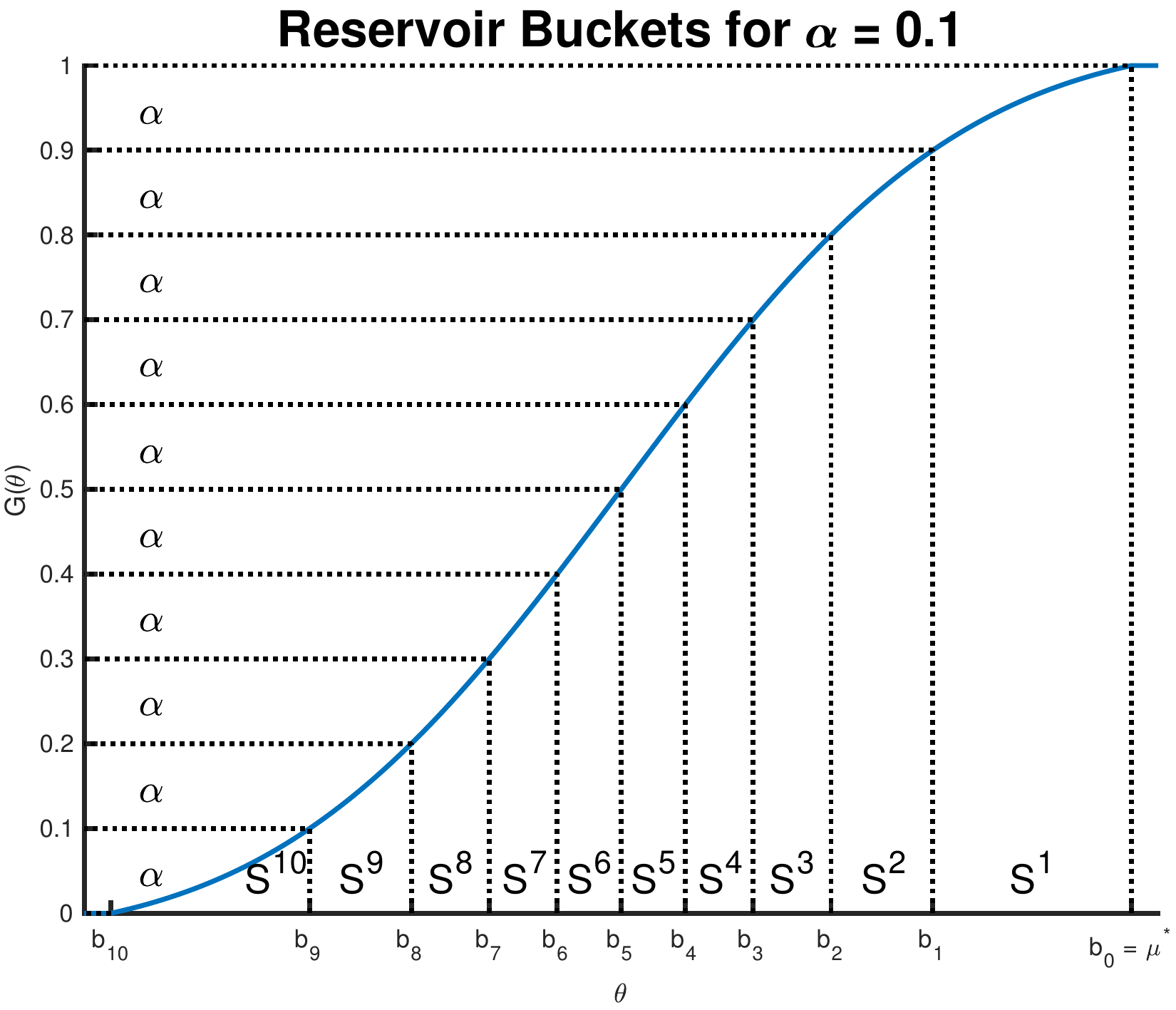}
	\caption{Our reservoir partition.
		Each consecutive $\alpha$-interval on the CDF defines some
		subset $\ArmGrp{i}$.}
	\label{fig-lb-buckets}
	\vspace{-30pt}
\end{wrapfigure}

It also relies on the following partition of the arms in $\Arms$ by their
expected rewards.
Let $m = \lceil 1/\alpha \rceil$. We partition $\Arms$ into subsets
$\ArmGrp{i}$ for $1 \le i \le m$,
where $\ArmGrp{i} = \{\arm \in \Arms:\ArmMean(\arm) \in ]b_i,b_{i-1}]\}$.
The interval boundaries $b_i$ are defined so that each subset has measure
$\alpha$ under the reservoir distribution $\ExpRewardDensity$,
with the possible exception of the subset with smallest expected reward.
In particular, $b_0 = \ArmMean^*$ and $b_i$ lies at the boundary between subsets
$i$ and $i+1$.
\begin{align}
b_i = \left\{ \begin{array}{ll}
    \ArmMean^* & \text{if $i = 0$} \\
    \ExpRewardInvCDF(G(b_{i-1}) - \alpha) & \text{if $i \ge 1$}
\end{array}\right.,
\end{align}
where $\ArmMean^*$ is defined in Eq.~\ref{eq-mustar}.
See Figure~\ref{fig-lb-buckets} for an illustration.

In the
Bernoulli case,
Assumption~\ref{ass:mu} reduces to $\ArmMean^* < 1$ (no arm has perfect performance);
when the set of possible
means $\ExpRewards$ is unbounded it always holds as $G$ has a finite
support.

\begin{assumption}\label{ass:mu} $\ArmMean^* < \sup_{\expReward \in \ExpRewards} \expReward$. 
 \end{assumption}

\begin{theorem}\label{thm-lb}
Fix some $\alpha, \delta \in ]0,1[$. Any $(\alpha,\delta)$-correct algorithm
needs an expected sample
complexity $\ETrueModel{\tau}$ that is lower bounded as follows.
\begin{align*}
\ETrueModel{\tau} \ge & \left(
    \frac{1}{\ArmKL{\ArmMean^*}{b_2}}
  + \sum_{i=2}^{m-1} \frac{1}{
            \ArmKL{b_i}{\ArmMean^*}
     }
\right)
\log \frac{1}{2.4\delta}.
\end{align*}
\end{theorem}

\begin{remark}\label{rem-finite-lb}
When $\Arms$ is finite s.t. $|\Arms|=K$, if we choose a uniform
reservoir and let
$\alpha = 1/K$, then $|\ArmGrp{i}| = 1$ for all $i$ and our lower bound
reduces to the bound obtained by \cite{JMLR15}
for best arm identification with $\epsilon=0$. Assuming arm means $\expReward_1 > \expReward_2 \geq \dots
\geq \expReward_K $, one has
\begin{align*}
\ETrueModel{T} \ge &\left[ \frac{1}{\ArmKL{\expReward_1}{\expReward_2}} +
\sum_{i=2}^K\frac{1}{\ArmKL{\expReward_i}{\expReward_1}}\right] \log
\frac{1}{2.4\delta}.
\end{align*}
\end{remark}

\subsection{Proof of Theorem~\ref{thm-lb}}

The proof relies on the following lemma that expresses a change of
measure in an infinite bandit model. Its proof is detailed in
Appendix~\ref{supp-proofs}. 

\begin{lemma}\label{lem-CD} Let $\ArmMeanAlt : \Arms \rightarrow \ExpRewards$ be an
alternative mean-mapping.
Let
$T_i(t) = \sum_{s=1}^t\1{\arm_s \in \ArmGrp{i}}$
be the number of times an arm in $\ArmGrp{i}$ has been
 selected.
For any stopping time $\sigma$ and
any
event $C \in H_\sigma$,
\vspace{-15pt}
\begin{align*}
\sum_{i=1}^m \ETrueModel{T_i(\sigma)} \sup_{\arm \in \ArmGrp{i}} \ArmKL{\ArmMean(\arm)}{\ArmMeanAlt(\arm)}
\geq  \BernKL{\PrTrueModel{C}}{\PrAltModel{C}},
\end{align*}
where $\BernKL{x}{y}=x\log(x/y) + (1-x)\log((1-x)/(1-y))$ is the Bernoulli relative
entropy.
\end{lemma}

Let $\tau^i = T_i(\tau)$ be the (random) number of draws from arms in 
$\ArmGrp{i}$,
so $\tau = \sum_{i=1}^m \tau^i$. Our lower bound on $\ETrueModel{\tau}$ follows
from
bounds on each of the $\ETrueModel{\tau^i}$.
We omit $\ArmGrp{m}$ because its measure may be less than $\alpha$.
By Assumption~\ref{ass:mu}, there is $\epsilon>0$ such that $\ArmMean^* +
\epsilon < \sup_{\expReward\in \ExpRewards} \expReward$.
Fix $i$ between $2$ and $m-1$ and define an alternative arm reward mapping
$\ArmMeanAlt^i(\arm)$ as follows.
\begin{align}
\ArmMeanAlt^i(\arm) = \left\{
\begin{array}{ll}
\ArmMean^* + \epsilon & \text{if } \arm \in \ArmGrp{i} \\
\ArmMean(\arm) & \text{otherwise}
\end{array}
\right.
\end{align}

This
 mapping induces an alternative reservoir distribution $g^i$
under which for all $i < m$, 
$\GTargetNoEpsAlt = \ArmGrp{i}$ because
$\ArmGrp{i}$ has measure $\alpha$
(as $\ArmMeasure$ is unchanged)
and under $\ArmMeanAlt^i$ the expected rewards of its arms
are above all other arms by at least $\epsilon$.
Also, by construction $\GTargetNoEps = \ArmGrp{1}$.

Define the event $C_\ArmMean = (\hat{s}_\tau \in \ArmGrp{1})$. Any
$(\alpha,\delta)$-correct algorithm thus satisfies
$\PrTrueModel{ C_\ArmMean } \geq 1 - \delta$
and $\PrAltModel{ C_\ArmMean } \leq \delta$. 
First using some monotonicity properties of the binary
relative entropy $\BernKL{x}{y}$, one has 
$
\BernKL{\PrTrueModel{C_\ArmMean}}{\PrAltModelI{C_\ArmMean}}
	\ge \BernKL{1-\delta}{\delta}
    \ge \log\frac{1}{2.4\delta},
$
where the second inequality is due to \cite{JMLR15}.

Applying Lemma~\ref{lem-CD} to event $C_\ArmMean$ and using the fact that
$\ArmMeanAlt^j(\arm)=\ArmMean(\arm)$ for all
$j\neq i$, one obtains
%
$
 \ETrueModel{T^i}\sup_{\arm \in
\ArmGrp{i}} \ArmKL{\ArmMean(\arm)}{\ArmMean^*+\epsilon} \geq \log\frac{1}{2.4\delta}.
$
%
Letting $\epsilon$ go to $0$ yields, for all $i\neq 1$,
\begin{align*}
\ETrueModel{\tau^i} \ge \frac{1}{\sup_{\arm \in \ArmGrp{i}}
        \ArmKL{\ArmMean(\arm)}{\ArmMean^*}}
        \log\left(\frac{1}{2.4\delta}\right),
\end{align*}
and $\sup_{\arm \in \ArmGrp{i}}
\ArmKL{\ArmMean(\arm)}{\ArmMean^*} \leq \ArmKL{b_i}{\ArmMean^*}$ as $\expReward \mapsto \ArmKL{\expReward}{\ArmMean^*}$ is
decreasing when $\expReward < \ArmMean^*$.

We now define the alternative mean rewards mapping, for $\epsilon>0$ small
enough
\begin{align}
\ArmMeanAlt^1(\arm) = \left\{
\begin{array}{ll}
b_2 - \epsilon & \text{if } \arm \in \ArmGrp{1} \\
\ArmMean(\arm) & \text{otherwise}
\end{array}
\right.
\end{align}
One has $\GTargetNoEps=\ArmGrpEmp{1}$ whereas
$\GTargetNoEpsAltOne=\ArmGrpEmp{2}$, hence
letting $C_\ArmMean = (\hat{s}_\tau \in \ArmGrpEmp{1})$ satisfies
$\PrTrueModel{C_\ArmMean} \geq 1-\delta$ and $\PrAltModelOne{C_\ArmMean} \leq \delta$.
Using the same reasoning as before yields
\begin{align*}
\log\left(\frac{1}{2.4\delta}\right) \leq \ETrueModel{\tau^1} \sup_{\arm \in \ArmGrp{1}}
\ArmKL{\ArmMean(\arm)}{b_2-\epsilon}
= \ArmKL{\ArmMean^*}{b_2-\epsilon}
\end{align*}
Letting $\epsilon$ go to zero yields 
$\ETrueModel{\tau^1} \ge \displaystyle\frac{1}{\ArmKL{\ArmMean^*}{b_2}}\log\left(\frac{1}{2.4\delta}\right)$.
\qed

One can prove an $\epsilon$-relaxed version of this
theorem, which provides a
lower bound on the number of samples needed to
find an arm whose expected reward is within $\epsilon$ of the top-$\alpha$ 
fraction of arms with probability at least $1-\delta$.
When $\epsilon>0$ multiple subsets may contain such arms, and the 
proof approach above does not work for these subsets.
We instead adopt the strategy of
\cite{Mannor04thesample}:
at most one such subset can have 
probability greater than $1/2$ of its arms being chosen by the algorithm,
so we exclude this subset from our bound.
We arrive at the following, which holds when $\ArmMean^*+\epsilon$ is in $\ExpRewards$ (i.e. for $\epsilon$ small enough).
\begin{remark}
Fix some $\alpha, \epsilon, \delta \in ]0,1[$,
and let $q$ be the number of subsets containing arms within $\epsilon$ of the top $\alpha$ fraction.
Any $(\alpha,\epsilon,\delta)$-correct algorithm needs an expected sample
complexity $\ETrueModel{\tau}$ that is lower bounded as follows.
\vspace{-5pt}
\begin{align*}
\ETrueModel{\tau} \ge & \left(
    \frac{q - 1}{\ArmKL{b_1-\epsilon}{\ArmMean^*+\epsilon}}
  + \sum_{i=q + 1}^{m-1} \frac{1}{
            \ArmKL{b_i}{\ArmMean^*+\epsilon}
     }
\right)
\log \frac{1}{4\delta}.
\end{align*}
\end{remark}

\section{Algorithm and Upper Bound}\label{sec-fc-upper}

In this section we assume that $P_{\ExpRewards} = \Set{ p_{\expReward}, \expReward \in \ExpRewards }$ is a one-parameter exponential family, meaning that there exists some twice differentiable convex function $b(\expReward)$ and some reference measure $\nu$ such that $p_\expReward$ has a density $f_\expReward$ with respect to $\nu$, where 
$
f_\expReward(x) = \exp(\expReward x - b(\expReward)).
$
Distributions in an exponential family can indeed be parameterized by their means as $\ArmMean=b^{-1}(\expReward)$.
We do not make any new assumptions on the reservoir distribution.

Under these assumptions on the arms, we present and analyze 
a two-phase algorithm called
\texttt{$(\alpha,\epsilon)$-KL-LUCB}.
We prove the $(\alpha,\epsilon,\delta)$-correctness of this algorithm and a high
probability upper bound on its sample complexity in terms of the complexity of
the reservoir distribution induced by the arm measure $\ArmMeasure$ and the
arm reward mapping function $\ArmMean$.
We also show how to
obtain $(\alpha,\delta)$-correctness
under assumptions on the tail of the reservoir.

\subsection{The algorithm}\label{kl-lucb}

\begin{wrapfigure}{r}{0.55\textwidth}
\vspace{-35pt}
\begin{minipage}{0.55\textwidth}
\begin{algorithm}[H]
\caption{$(\alpha,\epsilon)$-KL-LUCB}\label{alg-alpha-kl-lucb}
\begin{algorithmic}
\REQUIRE{$\alpha, \epsilon, \delta > 0$}
\STATE{$n = \frac{1}{\alpha} \ln\frac{2}{\delta}$}
\FOR{$a \leftarrow 1$ \textbf{to} $n$}
    \STATE{draw arm $\arm^a \sim \ArmMeasure$}
    \STATE{sample arm $\arm^a$ once}
\ENDFOR
\STATE{$t = n$ (current number of samples drawn)}
\STATE{$B(n) = \infty$ (stopping index)}
\STATE{Compute confidence bounds $U_a(n)$ and $L_a(n)$}
\WHILE{$B(t) > \epsilon$}
	\STATE{Draw arm $\arm^{\hat{a}(t)}$ and $\arm^{\hat{b}(t)}$}
    \STATE{$t = t + 2$}
    \STATE{Update confidence bounds, compute $\hat{a}(t)$ and $\hat{b}(t)$}
	\STATE{$B(t) = U_{\hat{b}(t)}(t) - L_{\hat{a}(t)}(t)$}
\ENDWHILE
\STATE \textbf{return} {$w^{\hat{a}(t)}$}
\end{algorithmic}
\end{algorithm}
\end{minipage}
\vspace{-5pt}
\end{wrapfigure}

\texttt{$(\alpha,\epsilon)$-KL-LUCB}, presented as Algorithm~\ref{alg-alpha-kl-lucb}, is a two-phase algorithm. It first queries $n=\frac{1}{\alpha} \log \frac{2}{\delta}$ arms
$\arm^1,\dots,\arm^n$ from $\ArmMeasure$, the measure over $\Arms$, and then runs the \texttt{KL-LUCB} algorithm \citep{COLT13} on the queried arms.
\texttt{KL-LUCB} identifies the $m$-best arms in a multi-armed bandit model, up to some $\epsilon>0$. We use it with $m=1$. 
This algorithm adaptively selects pairs of arms to sample from based on confidence intervals on the means of the arms. These confidence intervals rely on some 
exploration rate
\vspace{-5pt}
\begin{align}\label{explo-rate}
	\beta(t,\delta) := \log ( k_1 n t^\gamma / \delta ),
\end{align}
for
constants $\gamma>1$ and $k_1 > 2(1+\frac{1}{\gamma}-1)$. The upper and lower confidence bounds are
\begin{align} 
U_a(t) &:= \max \Set{ \expReward \in \ExpRewards : N_a(t) \ArmKL{\hat{p}_a(t)}{\expReward} \le \beta(t, \delta) }\label{kl-ub}\\
L_a(t) &:= \min \Set{ \expReward \in \ExpRewards : N_a(t) \ArmKL{\hat{p}_a(t)}{\expReward} \le \beta(t, \delta) }\label{kl-lb},
\end{align}
where $N_a(t) = \sum_{s=1}^t \1{\armRV_s = w^a}$ is the number of times arm $\arm^a$ was sampled by round $t$ and 
$\hat{p}_a(t) = \frac{1}{N_a(t)}\sum_{s=1}^t \1{\armRV_s = w^a}\rewardRV_s$
is the empirical mean reward of arm $\arm^a$ at round $t$, where $Z_s$ is an i.i.d. draw from arm $W_s$,
with distribution $p_{\ArmMean(W_s)}$.
Recall that $\ArmKL{\ArmMean_1}{\ArmMean_2}$ is the KL divergence between
arm distributions parameterized by their means $\ArmMean_1$ and $\ArmMean_2$.

For each queried arm $\arm^a$,
the algorithm maintains a confidence interval
$\ArmInterval = [L_a(t),U_a(t)]$
on $\ArmMean(w^a)$, and at any even round $t$ selects
two arm indexes: 
 (1) the empirical best arm $\hat{a}(t) \in {\argmax}_{a=1,\dots,n} \ \hat{p}_a(t)$, and
(2) the arm among the empirical worst arms that is most likely to be mistaken with $\hat{a}(t)$, $\hat{b}(t) = {\argmax}_{a \neq \hat{a}(t)} \ U_a(t)$.
The two arms are sampled: $\armRV_t = w^{\hat{a}(t)}$ and $\armRV_{t+1}=w^{\hat{b}(t)}$ and the confidence intervals are updated. 
The algorithm terminates when the overlap between the associated confidence intervals is smaller than some $\epsilon>0$: 
$
\tau = \inf\Set{t \in \N^* : L_{\hat{a}(t)}(t) > \max_{a \neq \hat{a}(t)} U_a(t) + \epsilon }.
$
The recommendation rule is $\hat{s}_\tau = w^{\hat{a}(t)}$.

%
%
\subsection{$(\alpha,\epsilon,\delta)$-Correctness}\label{sec-correctness}

For the algorithm to be $(\alpha,\epsilon,\delta)$-correct, it is sufficient that the following two events occur: 
\begin{itemize}[nosep]
    \item $A$ is the event that some $w^a$ was drawn from the
         top-$\alpha$ fraction of the reservoir.
    \item $B$ is the event that the \texttt{KL-LUCB} algorithm succeeds in 
identifying an arm within $\epsilon$ of the best arm among the $n$ arms drawn in the initialization
phase.
\end{itemize}
Indeed, on $A\cap B$ the recommended arm $\hat{s}_\tau = w^{\hat{a}(\tau)}$ satisfies 
$
\ArmMean(\hat{s}_\tau)  >  \max_{a} \ArmMean(w^a) - \epsilon  \geq  \ExpRewardInvCDF(1-\alpha) - \epsilon, 
$
hence $\hat{s}_\tau$ belongs to the top-$\alpha$ fraction, up to $\epsilon$. We prove in
Appendix~\ref{proof:correctness}
that $\Pr{A^\complement} \leq \delta/2$ and $\Pr{B^\complement}\leq \delta/2$,
which yields the following result. 
\vspace{-2pt}
\begin{lemma}\label{correctness}
With $\beta(t,\delta)$ defined in \eqref{explo-rate}, \texttt{$(\alpha,\epsilon)$-KL-LUCB} returns an arm from $\GTargetEps$
with probability at least $1-\delta$.
\end{lemma}
\vspace{-2pt}
It follows that when the parameter $\epsilon$ is chosen small
enough, e.g.,
$
\epsilon < \ExpRewardInvCDF\left(1 - \frac{\alpha}{2}\right) - \ExpRewardInvCDF\left(1 - {\alpha}\right),
$
\texttt{$(\alpha,\epsilon)$-KL-LUCB} is $(\alpha/2,\delta)$-correct.
For example, under the tail assumption~\ref{ass:smoothness}, $\epsilon$ can be 
chosen of order $c\alpha^{1/\beta}$.
However, when nothing is known about the reservoir distribution (e.g. it may
not even be smooth) we are not aware of an algorithm to choose $\epsilon$
to provide a $(\alpha,\delta)$-correctness guarantee.

\subsection{Sample Complexity of \texttt{$(\alpha,\epsilon)$-KL-LUCB}}

Recall the partition of $\Arms$ into subsets $\ArmGrp{i}$ of measure
$\alpha$ for $1 \le i \le m$,
where $\ArmGrp{i} = \Set{\arm \in \Arms:\ArmMean(\arm) \in ]b_i,b_{i-1}]}$.
We define our sample complexity bound in terms of the complexity
term:

\vspace{-0.5cm}

\begin{align}
\overline{H}_{\alpha,\epsilon} =
    \frac{ 2 }{ \epsilon^2 } +
    \sum_{ i = 2 }^{m}\frac{1}{
    \max(\epsilon^2/2,\ArmCI{b_{i-1}}{b_1})},
\end{align}

\vspace{-0.3cm}

where $\ArmCI{p}{q}$ is the Chernoff information
between two reward distributions parameterized by their means $p$ and $q$. This 
quantity is closely related to the KL-divergence: it is defined as 
$\ArmCI{p}{q}=\ArmKL{z^*}{p}$ where $z^*$ is the unique solution in $z$ to $\ArmKL{z}{p} = \ArmKL{z}{q}$. 

Let the random variable $\tau$ be the number of samples used by
\texttt{$(\alpha,\epsilon)$-KL-LUCB}. The following upper bound on $\tau$ holds.

\begin{theorem}\label{samplecomp}
Let $\alpha,\delta$ such that $0<\delta \leq \alpha \leq 1/3$.
The \texttt{$(\alpha,\epsilon)$-KL-LUCB} algorithm with exploration rate
$\beta(t,\delta)$ defined by \eqref{explo-rate} and a parameter $\epsilon>0$
is $(\alpha,\epsilon,\delta)$-correct and satisfies, with probability at least $1-7\delta$,
\vspace{-10pt}
\begin{align*}
    \tau \le 12 C_0(\gamma) \overline{H}_{\alpha,\epsilon}
    \log^2 \frac{1}{\delta} + o\left( \log^2 \frac{1}{\delta} \right),
\end{align*}
%
%
with $C_0(\gamma)$ such that
$C_0(\gamma) \ge \gamma \log(C_0(\gamma)) + 1 + \frac{\gamma}{e}$.
\end{theorem}

Theorem~\ref{samplecomp} only presents the leading term in $\delta$ (when $\delta$ goes to zero) of the sample complexity upper bound, but an explicit upper bound can be extracted from the proof of Theorem~\ref{samplecomp}, given in Appendix~\ref{sec-proof}.  

\subsection{Comparison and Discussion}
\label{sec-compare-lb-ub}

Our $\epsilon$-relaxed bounds simplify,
for appropriate constants $c_1,c_2$ and small enough $\epsilon$, to
\begin{align*}
\ETrueModel{ \tau } & \ge c_1 \left(
	\frac{1}{\ArmKL{b_1-\epsilon}{b_0+\epsilon}}
	+ \sum_{i=3}^{m-1} \frac{1}{\ArmKL{b_i}{b_0 + \epsilon}}
\right) \log \frac{1}{\delta},
\\
\tau &\le
c_2 \left( 
  \frac{4}{\epsilon^2}
  + \sum_{i=2}^{m-1} \frac{1}{\ArmCI{b_{i}}{b_1}}
\right) \log^2 \frac{1}{\delta}.
\end{align*}
A log factor separates our bounds,
and the upper bound complexity term is slightly
larger than that in the lower bound.
KL-divergence in the lower bound is of comparable scale to Chernoff information in the upper bound:  in the Bernoulli case one has 
$
\frac{(\ArmMean^*-x)^2}{2} < \ArmCI{x}{\ArmMean^*}
	< \ArmKL{x}{\ArmMean^*}
	< \frac{(\ArmMean^* - x)^2}{\ArmMean^*(1-\ArmMean^*)}.
$
However, for $i\neq 1$, $\ArmCI{b_i}{b_1}$ is slightly smaller than $\ArmKL{b_i}{b_0 + \epsilon}$,
while $\epsilon^2$ is smaller than $\ArmKL{b_1-\epsilon}{b_0+\epsilon}$. 
These differences are reduced as $\alpha$ is decreased.

When $\delta$ is not too small, the extra $\log \frac{1}{\delta}$ factor is small compared to the constants in our upper bound.
It is well-established 
for finite bandit models that for a wide variety of
algorithms the sample complexity 
scales like
$\log \frac{1}{\delta}$. The additional log factor comes from the fact that each phase of our algorithm needs a $\delta$-correctness
guarantee.
In our first phase, we choose a number of arms to draw from the reservoir
without drawing any rewards from those arms.
In the second phase, we observe rewards from our arms without drawing any new
arms from the reservoir.
It is an interesting open question to prove whether in any such two-phase algorithm the
$\log^2 \frac{1}{\delta}$ term is avoidable.
It is not hard to show that the first phase must draw at least $\frac{c}{\alpha} \log \frac{1}{\delta}$ arms for some constant $c$ in order to obtain a single arm from the top-$\alpha$ fraction with high probability, but the \emph{expected} number of arms in the top-$\alpha$ fraction is already $c \log \frac{1}{\delta}$ in this case.
The second phase can be reduced to a problem of finding one of the top $m$ arms, or of finding one arm above the unknown threshold $\ExpRewardInvCDF(1 - \alpha)$, but we are not aware of a lower bound on these problems even for the finite case.

Despite all this, it seems likely that a one-phase algorithm can avoid the
quadratic dependence on $\log \frac{1}{\delta}$.
Indeed, \cite{Jamieson:2016wd}
provides such an algorithm for the special case of reservoirs involving just
two expected rewards.
They employ a subroutine which 
returns the target coin with constant probability
by drawing a number of coins that does not depend on $\delta$.
They wrap this subroutine in a $\delta$-correct algorithm which iteratively
considers progressively more challenging reservoirs,
terminating when
a target coin is identified.
We agree with the authors that adapting this approach for general reservoirs is
an interesting research direction.
However their method relies on the special
shape of their reservoir
and it is not immediately clear how it might be generalized.

%

\section{Conclusion}\label{sec-conclusion}
In contrast with previous approaches to bandit models, we have
limited consideration to changes of distribution which change only the mean mapping 
$\ArmMean$ and not the measure $\ArmMeasure$ over arms $\Arms$.
This allows us to analyze infinite bandit models without a need to integrate
over the full reservoir distribution, so we can prove results for reservoirs
which are not even smooth.
We proved a lower bound on the sample complexity of the problem,
and we introduced an algorithm with an upper bound within a log factor of our
lower bound.

An interesting future direction is to study improved algorithms, namely
one-phase algorithms which alternate between sampling arms to estimate the 
reservoir and drawing new arms to obtain better arms with higher confidence.
These algorithms might be able to have only a $\log\frac{1}{\delta}$ instead of
$\log^2 \frac{1}{\delta}$ dependency in the upper bound.
In practice, however, the algorithm we present exhibits good empirical 
performance.

\bibliographystyle{plain}

\small

\paragraph{Acknowledgement.} We thank Virgil Pavlu for the fruitful discussions and his efforts that made this project better. E. Kaufmann acknowledges the support of the French Agence Nationale de la Recherche (ANR), under grant \texttt{ANR-16-CE40-0002} (project BADASS). 

\bibliography{paper,biblioEmilie}

\normalsize

\newpage

\appendix 

\section{Changes of distribution in infinite bandit models}\label{supp-proofs}

\subsection{Proof of Lemma~\ref{lem-CD}} \label{proof-CD}

We describe in this section the key results that allow us to adapt
\emph{changes of distribution} arguments to the infinite bandit setting.
Lemma~\ref{lem-CD} follows easily from Lemma~\ref{up-2014arXiv1407.4443K} and 
Lemma~\ref{main-lb-lemma} that are stated below and proved in the next two sections.

All the regret or sample complexity lower bounds in bandit models rely on
change of distributions arguments (see, e.g.
\cite{LaiRobbins85bandits,BurnKat96,Bubeck10BestArm}). A change of distribution
relates the probability of an event under a given bandit model to the
probability of that event under an alternative bandit model, that is ``not too
far'' from the initial model but under which the performance of the algorithm
is supposed to be completely different. \cite{Combes14Lip,JMLR15} recently
found an elegant formulation for such a change of distribution in terms of the
expected log-likelihood ratio and we explain below how we generalize these
tools to the infinite bandit model.

Given an infinite bandit model $(\ArmMeasure,\ArmMean)$, one may consider an alternative
bandit model $(\ArmMeasure,\ArmMeanAlt)$ in which the measure $\ArmMeasure$ is similar but the
mean function is different: $\ArmMeanAlt \neq \ArmMean$. As mentioned in
Section~\ref{setup}, we consider strategies such that
$\PrTrueModel{ \armRV_t | \historyRV_{T-1} }$ is independent from $\ArmMean$. Hence, defining the log-likelihood
ratio between $(\ArmMeasure,\ArmMean)$ and $(\ArmMeasure,\ArmMeanAlt)$ at round $t$ as
$L_{\ArmMean,\ArmMeanAlt}(t) = \log(\ell(\historyRV_t ; \ArmMean,\ArmMeasure) / \ell(\historyRV_t; \ArmMeanAlt,\ArmMeasure))$, where the likelihood is defined in 
\eqref{Likelihood}, one has 
\begin{align}
L_{\ArmMean,\ArmMeanAlt}(T)
    &= \sum_{t=1}^T \log \frac{f_{\ArmMean(\armRV_t)}(\rewardRV_t)}{f_{\ArmMeanAlt(\armRV_t)}(\rewardRV_t)}.\label{logLikelihood}
\end{align}

The following result generalizes Lemma 1 in \cite{JMLR15} to the infinite
bandit model. It permits to relate the expected log-likelihood ratio to the
probability of any event under the two different models.

\begin{lemma}\label{up-2014arXiv1407.4443K}
\label{lem-log-lik}
Let $\sigma$ be a stopping time and $\ArmMean$ and $\ArmMeanAlt$ be two reward mappings.
For any event $C$ in $H_\sigma$,
\begin{align*}
\ETrueModel{ L_{\ArmMean,\ArmMeanAlt}(\sigma) }
	\ge \BernKL{\PrTrueModel{C}}{\PrAltModel{C}},
\end{align*}
where $\BernKL{x}{y}=x\log(x/y) + (1-x)\log((1-x)/(1-y))$ is the Bernoulli relative
entropy.
\end{lemma}

The next result provides an upper bound on the expected log-likelihood ratio.
While in a classic multi-armed bandit model, the log-likelihood can be
expressed as a sum that features the expected number of draws of each arm, such
a quantity would not be defined in the infinite bandit model. Hence, we need to
introduce a partition of $\Arms$.

\begin{lemma}\label{main-lb-lemma}
Fix $\ArmGrpEmp{1},\dots,\ArmGrpEmp{m}$ a partition of $\Arms$ and let
$T_i(t) = \sum_{s=1}^t\1{\arm_s \in \ArmGrp{i}}$
be the number of times an arm in $\ArmGrp{i}$ has been selected.
For any stopping time $\sigma$,
\begin{align*}
\ETrueModel{ L_{\ArmMean,\ArmMeanAlt}(\sigma) }
\leq \sum_{i=1}^m \ETrueModel{ T_i(\sigma) } \sup_{\arm \in \ArmGrp{i}} \ArmKL{\ArmMean(\arm)}{\ArmMeanAlt(\arm)}.
\end{align*}
\end{lemma}

\subsection{Proof of Lemma~\ref{up-2014arXiv1407.4443K}}

The proof for the infinite case follows the argument by \cite{JMLR15} for the
finite case. First, the conditional Jensen's inequality is applied, given the
convexity of
$\exp(-x)$.
The expectation derivations hold for our infinite arms case without
modification.
The only necessary statement for which the finite case proof needs updating,
$\PrAltModel{C} = \ETrueModel{ \1{C} \exp(-L_{\ArmMean,\ArmMeanAlt}(\sigma)) }$,
is proven for our infinite case setting as Lemma~\ref{lem-pr-exp}.

\begin{lemma}\label{lem-pr-exp}
Let $\ArmMean$ and $\ArmMeanAlt$ be two arm reward mappings,
$L_{\ArmMean,\ArmMeanAlt}(T)$ be the log likelihood ratio defined in \eqref{logLikelihood}
and $C$ be an (event) subset of histories of length $T$.
Then
\begin{align*}
\PrAltModel{C}
	= \ETrueModel{ \1{C} \cdot \exp(-L_{\ArmMean,\ArmMeanAlt}(T)) }
\end{align*}
\end{lemma}

\paragraph{Proof of Lemma~\ref{lem-pr-exp}.}
Let $\ell(\historyRV_T ;\ArmMean,\ArmMeasure)$ be the likelihood function defined in \eqref{Likelihood}.
We introduce furthermore the notation $h_t = (\armRV_1,\rewardRV_1,\dots,\armRV_t,\rewardRV_t)$, $\bm \arm_T=(\armRV_1,\dots,\armRV_T)$ and $\bm z_T =(\rewardRV_1,\dots,\rewardRV_T)$. Recall that a strategy is such that conditional density of $\armRV_t$ given $\historyRV_{t-1}$ does not depend on the mean reward mapping but only on the reservoir distribution: we denote it by $\PrNone_\ArmMeasure(\armRV_t | \historyRV_{t-1})$.
The proof of Lemma~\ref{lem-pr-exp} follows from the following inequalities. 
\begin{align*}
\PrAltModel{C} =
  \EAltModel{ \1{C} }
 & =  \int \1{C}(\historyRV_T) \ell(\historyRV_T; \ArmMeanAlt, \ArmMeasure) \diff \historyRV_T \\
 & = \int \1{C}(h_T)
 \prod_{t=1}^Tf_{\ArmMeanAlt(\armRV_t)}(\rewardRV_t)
 \PrNone_\ArmMeasure(\armRV_t | h_{t-1}) \diff \bm \arm_T  \diff \bm z_T \\
 & = \int \1{C}(h_T)
 	\prod_{t=1}^T\frac{f_{\ArmMeanAlt(\armRV_t)}(\rewardRV_t)}{f_{\ArmMean(\armRV_t)}(\rewardRV_t)} \prod_{t=1}^T f_{\ArmMean(\armRV_t)}(\rewardRV_t)
 \PrNone_{\ArmMeasure}(\armRV_t | h_{t-1}) \diff \bm \arm_T \diff \bm z_T \\
  & = \int \1{C}(H_T) \prod_{t=1}^T\frac{f_{\ArmMeanAlt(\armRV_t)}(\rewardRV_t)}{f_{\ArmMean(\armRV_t)}(\rewardRV_t)} \ell(\historyRV_T;\ArmMean,\ArmMeasure) \diff \historyRV_T  \\
  & = \ETrueModel{
 \1{C} \cdot\prod_{t=1}^{T} \frac{ f_{\ArmMeanAlt(\armRV_t)}(\rewardRV_t) }{ f_{\ArmMean(\armRV_t)}(\rewardRV_t) } }.
\end{align*}
\qed

\subsection{Proof of Lemma~\ref{main-lb-lemma}}

The proof follows from the following inequalities.
\begin{align*}
& \ETrueModel{ L_{\ArmMean,\ArmMeanAlt}(\sigma) } \\
&= \ETrueModel{ \sum_{t=1}^\infty
	\ETrueModel{ \left.\1{\sigma \geq t-1} \log
	\frac{f_{\ArmMean(\armRV_t)}(\rewardRV_t)}{f_{\ArmMeanAlt(\armRV_t)}(\rewardRV_t)} \right|
	\historyRV_{t-1}}}\\
& = \ETrueModel{ \sum_{t=1}^\infty \1{\sigma \geq t-1}
\ArmKL{\ArmMean(\armRV_t)}{\ArmMeanAlt(\armRV_t)}} \\
& \leq  \ETrueModel{ \sum_{i=1}^m \sum_{t=1}^\sigma \1{\armRV_t \in \ArmGrp{i}}
\sup_{\arm \in \ArmGrp{i}} \ArmKL{\ArmMean(\arm)}{\ArmMeanAlt(\arm)} }
\end{align*}

\section{Proof of Lemma~\ref{correctness}}\label{proof:correctness}

Letting
\begin{align*}
A & =  \left(\exists a \leq n : \ArmMean(w^a) > \ExpRewardInvCDF(1-\alpha)\right) \\
B & =  \left(\ArmMean_{\hat{a}(\tau)} \geq \max_{a} \ArmMean_a - \epsilon\right),
\end{align*}
Lemma~\ref{correctness} follows from the fact that $\Pr{A^\complement}\leq \frac{\delta}{2}$ and $\Pr{B^\complement} \leq \frac{\delta}{2}$, that we now prove. 

First, by definition of the reservoir distribution $G$ and the fact that $\ArmMean(\armRV^t)$ are i.i.d. samples from it, one has

\begin{align*}
 \Pr{A^\complement} & =  \Pr{\bigcap_{a = 1}^n\left( \ArmMean(\arm^a) \leq \ExpRewardInvCDF(1-\alpha)\right)}
 	= \left(1 - \alpha \right)^n \\
 & =  \exp\left(-n\log\left(\frac{1}{1- \alpha}\right)\right) \\
 & =  \exp\left(-\log\left(\frac{2}{\delta}\right)\frac{1}{\alpha}\log\left(\frac{1}{1- \alpha}\right)\right) \leq \frac{\delta}{2},
\end{align*}
using that $-\log(1-x)>x$. 

The upper bound on $\Pr{B^\complement}$ follows the same lines as that of the correctness of the \texttt{KL-LUCB} algorithm \citep{COLT13}, however note that we are able to use a smaller exploration rate compared to this work.
	Abusing notation slightly we let $\ArmMean_a := \ArmMean(\arm^a)$,
	and letting $(1) = \argmax_a \ArmMean_a$ we have
\begin{align*}
 B^\complement & \subseteq  (\exists t \in \N : \ArmMean_{(1)} > U_{(1)}(t)) \!\!\!\!\!\bigcup_{a : \ArmMean_a < \ArmMean_{(1)} - \epsilon} \!\!\!\!\!\!\!\! (\exists t \in \N: L_a(t) > \ArmMean_a).
\end{align*}
Let $\operatorname{d}^-(x,y) := \ArmKL{x}{y}\1{x > y}$.
For each $a$, $\Pr{\exists t\in \N: L_a(t) > \ArmMean_a | \ArmMean_a}$ is upper bounded by
\begin{align*}
&\Pr{ \exists t \in \N : N_a(t) \operatorname{d}^-(\hat{p}_a(t),\ArmMean_a) > \beta(t,\delta) } \\
& \leq   \Pr{ \exists t \in \N : N_a(t) \operatorname{d}^-(\hat{p}_a(t),\ArmMean_a) > \beta(N_a(t),\delta) } \\
& \leq   \Pr{ \exists s \in \N : s \operatorname{d}^-(\hat{p}_a(t),\ArmMean_a) > \beta(s,\delta) } \\
& \leq  \sum_{s=1}^\infty \exp(-\beta(s,\delta)) \leq \frac{\delta}{k_1 n}\sum_{t=1}^\infty \frac{1}{t^\gamma} \leq \frac{\delta}{2n},
\end{align*}
where we use a union bound together with Chernoff's inequality and the fact that $k_1$ is chosen to be larger than $2 \sum_{t}\frac{1}{t^\gamma}$. Similar reasoning shows that
\begin{align*}
\Pr{\exists t \in \N : \ArmMean_{(1)} > U_{(1)}(t)}
	\leq \frac{\delta}{2n},
\end{align*}
and a union bound yields $\Pr{B^\complement} \leq \frac{\delta}{2}$.

\section{Proof of Lemma~\ref{lem-armNumAppBandit}}\label{proof:armNumAppBandit}

For every $i$, let $\cR_i$ be the set of arms $a \in \{1,\dots,n\}$ such that $w^a \in \ArmGrp{i}$. Letting $Y^i_a = \1{w^a \in \ArmGrp{i}}$, as $\ArmMeasure(\ArmGrp{i}) = \alpha$, $(Y^i_a)_a$ is i.i.d. with a Bernoulli distribution of parameter $\alpha$ and 
$|\cR_i| = \sum_{a=1}^n Y^i_a$. 

Using Chernoff's inequality for Bernoulli random variables yields 
\begin{align*}
 &\Pr{ |\cR_i| > 6 \log(1/\delta) }
 	= \Pr{ |\cR_i| > 3\alpha n } \\
 & =  \Pr{ \frac{1}{n}\sum_{a=1}^n Y^i_a > 3\alpha }
 	\leq \exp\left\{-n\BernKL{3\alpha}{\alpha}\right\},
\end{align*}
where $\BernKL{x}{y}=x\log(x/y) + (1-x)\log((1-x)/(1-y))$ is the binary relative entropy. 

Using Lemma~\ref{lem-ineq-kl} below for $\beta=2$, if $\alpha \leq 1/3$ it holds that
\begin{align*}
\BernKL{3\alpha}{\alpha} \geq (3\log 3 - 2)\alpha > \alpha
\end{align*}
and, using again the definition of $n$,
\begin{align*}
 &\Pr{ |\cR_i| > 6 \log(1/\delta) }
 	\leq \exp\left\{- 2 \log(1/\delta)\right\}
 	= \delta^2.
\end{align*}
Using a union bound on the $m$ subsets $\ArmGrp{i}$, and the fact that $m \leq 1/\alpha$ and $\delta \leq \alpha$, one has 
\begin{align*}
\Pr{ C^\complement }
	\leq m \delta^2
	\leq \frac{\delta^2}{\alpha}
	\leq \delta,
\end{align*}
which concludes the proof.

\begin{lemma}\label{lem-ineq-kl} Let $\beta > -1$. For all $\alpha \leq \frac{1}{1+\beta}$, 
\begin{align*}
\BernKL{(1+\beta)\alpha}{\alpha}
	\geq ((1+\beta)\log(1+\beta) - \beta) \alpha.
\end{align*}
\end{lemma}

This inequality is optimal in the first order in the sense that its two members are equivalent when $\alpha$ goes to zero.

\paragraph{Proof of Lemma~\ref{lem-ineq-kl}.} By definition
\begin{align*}
 \BernKL{(1+\beta)\alpha}{\alpha}
 	=& (1+\beta)\alpha\log(1+\beta) \\
 & + (1-(1+\beta)\alpha)\log\left(\frac{1-(1+\beta)\alpha}{1-\alpha}\right) \\
 =& (1+\beta)\alpha\log(1+\beta) \\
  &  +  (1-(1+\beta)\alpha)\log\left(1 - \frac{\beta\alpha}{1-\alpha}\right) 
\end{align*}
Now, using the fact that $1 - (1-\beta)\alpha >0$ for $\alpha \leq 1/(1+\beta)$ and the following inequality 
\[\forall x > -1, \ \ \log(1+x) \geq \frac{x}{1+x}\]
one obtains 
\begin{align*}
	\BernKL{(1+\beta)\alpha}{\alpha}
		\geq & (1+\beta)\alpha\log(1+\beta)
\\
   &  +  (1-(1+\beta)\alpha)\frac{-\beta\alpha}{1-\alpha - \beta\alpha}
\\
  = & \alpha \left((1+\beta)\log(1+\beta) - \beta\right)
\end{align*}
\qed

\section{Proof of Theorem~\ref{samplecomp}} \label{sec-proof}

We let $\ArmMean_a := \ArmMean(\arm^a)$ and
denote by $\bm\ArmMean = (\ArmMean_1,\dots,\ArmMean_n)$ the means of the $n$ arms that have been queried, sorted in decreasing order. 

In addition to events $A$ and $B$ defined in Section~\ref{sec-correctness}, we introduce the event $C$ that for every subset $i \leq m$, at most $6\log\frac{1}{\delta}$
        arms belong to $\ArmGrp{i}$:
\begin{align}
C = \bigcap_{i = 1,\dots,m} \left\{|\{ a \leq n : w^a \in \ArmGrp{i}\}| \leq 6\log({1}/{\delta})\right\}
\end{align}
We prove the following in
Appendix~\ref{proof:armNumAppBandit}.

\begin{lemma}\label{lem-armNumAppBandit} If $\delta \leq \alpha \leq 1/3$,
$\Pr{ C } \geq 1-\delta$.
\end{lemma}

For all $t\in\N$ we introduce the event 
\begin{align}
W_t = \bigcap_{1\leq a \leq n} (L_a(t) \leq \ArmMean_a \leq U_a(t))
\end{align}
and define $W = \cap_{t\in\N^*} W_t$. By the same argument as the one used in the proof of Lemma~\ref{correctness} (see
Appendix~\ref{proof:correctness}),
one 
can show that $\Pr{ W } \geq 1-2\delta.$

Fix $c \in [\ArmMean_2,\ArmMean_1]$. Our analysis relies on the following crucial statement.

\begin{proposition}[\cite{COLT13}]\label{kauffprop}Let
$\tilde{\beta}_a(t) := \sqrt{ \frac{\beta(t,\delta)}{2 N_a(t)} }$. If $W_t$ holds and $(U_{\hat{b}(t)} - L_{\hat{a}(t)}>\epsilon)$ then there exists $a\in \{\hat{a}(t),\hat{b}(t)\}$ such that
\begin{align}
c \in \ArmInterval \text{ and } \tilde{\beta}_a(t) > {\epsilon}/{2}.
\end{align}
\end{proposition}
Fixing some integer $T \in \N$, we now upper bound $\tau$ on the event $\cE := A\cap B\cap C\cap W$.
\begin{align*}
& \min(\tau,T)   \le  \sum_{t=1}^T\1{\tau > t} = n + 2\sum_{\substack{t  \in n + 2\N\\t \leq T}}\1{\tau > t} \\
& \leq  n + 2\sum_{\substack{t  \in n + 2\N\\t \leq T}}\1{U_{\hat{b}(t)} - L_{\hat{a}(t)} > \epsilon} \\
& \leq  n + 2\sum_{\substack{t  \in n + 2\N\\t \leq T}}\1{\exists a \in \{\hat{a}(t),\hat{b}(t)\}: c \in \ArmInterval, \tilde{\beta}_a(t) > \epsilon/2},
\end{align*}
using Proposition~\ref{kauffprop} and the fact that W holds. To ease the notation, we let $\ArmGrpEmp{t} = \{\hat{a}(t),\hat{b}(t)\}$ be the set of drawn arms at round $t$. Letting $\EpsCheapArms=\Set{a : \ArmCI{\ArmMean_a}{c} < \epsilon^2/2}$ and noting that $\tilde{\beta}_a(t) > \epsilon/2 \iff N_a(t) <
\beta(t,\delta)/(\epsilon^2/2)$, 
\begin{align*}
& \min(\tau,T)   \le  n + 2\!\sum_{a \in \EpsCheapArms}\!\sum_{\substack{t  \in n + 2\N\\t \leq T}}\!\!\! \1{a \in \ArmGrpEmp{t}}\1{N_a(t) < \beta(T,\delta)/(\epsilon^2/2)}\\
&  \hspace{1cm}+ 2\sum_{a \in \EpsCheapArms^\complement} \sum_{\substack{t  \in n + 2\N\\t \leq T}} \1{a \in \ArmGrpEmp{t}}\1{N_a(t)\ArmKL{\hat{p}_a(t)}{c} \leq \beta(T,\delta))}
\end{align*}
Now defining the event
\begin{align*}
G_T^a = \!\!\!\!\!\bigcup_{\substack{t  \in n + 2\N\\t \leq T}}\!\! \left\{\!N_a(t) > \! \frac{\beta(T,\delta)}{\ArmCI{\ArmMean_a}{c}} , N_a(t)\ArmKL{\hat{p}_a(t)}{c} \! \leq\!  \beta(T,\delta) \!\right\}
\end{align*}
From Lemma 1 in \cite{COLT13},
\begin{align}
\Pr{ G_T^a | \bm\ArmMean }
	\leq \frac{1}{\ArmCI{\ArmMean_a}{c}}\exp(-\beta(T,\delta)).\label{boundE1}
\end{align}
Introducing $\cF_T = \cap_{a \in \EpsCheapArms^\complement} (G_T^a)^\complement$, one can further upper bound $\tau$ on $\cE \cap \cF_T$ as 
\begin{align*}
& \min(\tau,T)   \le  n + 2\!\sum_{a \in \EpsCheapArms}\!\sum_{\substack{t  \in n + 2\N\\t \leq T}}\!\!\!\! \1{a \in \ArmGrpEmp{t}}\1{N_a(t) < \beta(T,\delta)/(\epsilon^2/2)}\\
&  \hspace{1cm}+  2\sum_{a \in \EpsCheapArms^\complement} \sum_{\substack{t  \in n + 2\N\\t \leq T}} \1{a \in \ArmGrpEmp{t}}\1{N_a(t) < \beta(T,\delta)/\ArmCI{\ArmMean_a}{c}} \\
& \hspace{1cm} \le  n + 2\underbrace{\left[\sum_{a \in \EpsCheapArms} \frac{2}{\epsilon^2} + \sum_{a \in \EpsCheapArms^\complement}\frac{1}{\ArmCI{\ArmMean_a}{c}}\right]}_{=: H(\bm\ArmMean,c,\epsilon)}\beta(T,\delta).
\end{align*}
We now provide a deterministic upper bound on $H(\bm\ArmMean,c,\epsilon)$, by summing over the arms in the different subsets $\ArmGrp{i}$.
Let $q$ be the number of subsets containing arms within $\EpsCheapArms$.
We choose $c$ such that $c>\ExpRewardInvCDF(1-\alpha) = b_1$ (such a choice is possible as $\ArmMean_1 >\ExpRewardInvCDF(1-\alpha)$ as event $A$ holds) and note that
	$\EpsCheapArms \subseteq \cup_{i\le q} \ArmGrpEmp{i}$.
One has 
\begin{eqnarray*}
H(\bm \ArmMean,c,\epsilon)
	\!\!& \leq &\!\!\!\!\!
	\sum_{i=1}^q
	\sum_{a : \arm_a \in \ArmGrp{i}} \frac{2}{\epsilon^2}
	+
	\sum_{i=q+1}^m
	\sum_{a : \arm_a \in \ArmGrp{i}}
	\!\!\frac{1}{\ArmCI{\ArmMean_a}{\ExpRewardInvCDF(1-\alpha)}}  \\
& \leq &\!\! 6 \overline{H}_{\alpha,\epsilon}\log(1/\delta),
\end{eqnarray*}
using that event $C$ holds and each $\ArmGrp{i}$ contains at least $6\log(1/\delta)$ arms. Hence on $\cE \cap \cF_T$, 
\[\min(\tau,T)   \le n + 12 \overline{H}_{\alpha,\epsilon}\log(1/\delta) \beta(T,\delta).\]
Applying this to  $T=T^*$ where 
\[T^* := \inf\{ T \in \N : n + 12 \overline{H}_{\alpha,\epsilon}\log(1/\delta) \beta(T,\delta) \leq T\},\]
one obtains $\min(\tau,T) \leq T$, hence $\tau \leq T$. We proved that 
\begin{align*}
\Pr{ \tau \leq T^* }
	\leq 1 - 4\delta - \Pr{ \cF_{T^*}^\complement }.
\end{align*}

An upper bound on $T^*$ can be extracted from Appendix E of \cite{COLT13}: 
\begin{align*}
T^* \leq 12C_0(\gamma)\overline{H}_{\alpha,\epsilon}\log\frac{1}{\delta} \log\left(\frac{12k_1n(\log(1/\delta)^\gamma)\overline{H}_{\alpha,\epsilon}^\gamma}{\delta}\right) + n.
\end{align*}
An upper bound on $\Pr{ \cF_{T^*}^\complement }$ concludes the proof: 
\begin{align*}
 \Pr{ \cF_{T^*}^\complement }
 	&\leq  2\delta + \Pr{ \cF_{T^*}^\complement \cap A \cap C } \\
	&= 2\delta + \E{ \Pr{ \cF_{T^*}^\complement | \bm \ArmMean } \1{A \cap C} }
\end{align*}
Now, on $A\cap C$ (which is used for the second inequality), using \eqref{boundE1},
\begin{align*}
 \Pr{ \cF_{T^*}^\complement | \bm \ArmMean }
 	&\leq \sum_{a \in \EpsCheapArms^\complement}
 		\frac{1}{\ArmCI{\ArmMean_a}{c}}\exp(-\beta(T^*,\delta)) \\
	&\leq 6\overline{H}_{a,\epsilon}\log(1/\delta)\exp(-\beta(T^*,\delta)) \\
	&\leq \left(\frac{6\overline{H}_{a,\epsilon}\log(1/\delta)}{k_1 n (T^*)^\gamma}\right)  \delta \leq \delta,
\end{align*}
where the last inequality follows from the definition of $T^*$. Thus $\Pr{ \cF_{T^*}^\complement } \leq 3\delta$.

\section{Empirical Results}\label{sec-empirical}

We exhibit
\texttt{$(\alpha,\epsilon)$-KL-LUCB} for infinite models of Bernoulli arms with
various parameter values.
In order to meet our assumption that $\ArmMean^* < 1$,
we truncate our $Beta$ distributions to have support on the
interval $]0, 0.95]$.
We report the fraction of runs in which it fails to find an arm within $\epsilon$ of the top-$\alpha$
fraction, the mean simple regret observed, and the mean budget used.
We have also tried various minor modifications to this algorithm which appear to 
reduce the sample complexity without much affecting its success rate or simple regret.
These include drawing a sample only from the least-sampled of the two arms
chosen at each round
and updating confidence intervals only for those arms whose upper bounds overlap 
with the top arm.
However, we provide no performance guarantees for the algorithm with these 
modifications.
Note that our results are for the algorithm as stated in the paper, and not with 
any of these modifications.

\begin{table}[h]
\caption{Performance of \texttt{$(\alpha,\epsilon)$-KL-LUCB}.
Mean of 100 runs.
``Effective $\alpha$'' gives the measure of arms meeting the $(\alpha,\epsilon)$
criteria.
``Errors'' indicates the fraction of runs where the $\alpha$
objective was not achieved; it should be below $\delta$.
\\
The budget $T$ is impacted roughly linearly by $1/\alpha$ and quadratically 
by $1/\epsilon$; the same regret is achieved with different budgets based
on parameter selection.
}
\label{tbl-results}
\vskip 0.15in
\begin{center}
\begin{small}
\begin{sc}
\begin{tabular}{lccc|cccc}
\hline
Reservoir & $\alpha$ & $\epsilon$ & $\delta$ & Effective $\alpha$ & Errors & Simple Regret & $T$ \\
\hline
Beta(1,1) & 0.025 & 0.024 & 0.05 & 0.049 & 0.00 & 0.008 & 51k \\
Beta(1,1) & 0.025 & 0.024 & 0.10 & 0.049 & 0.01 & 0.011 & 46k \\
Beta(1,1) & 0.050 & 0.010 & 0.05 & 0.060 & 0.02 & 0.015 & 113k \\
Beta(1,1) & 0.050 & 0.010 & 0.10 & 0.060 & 0.06 & 0.020 & 90k \\
Beta(1,1) & 0.050 & 0.048 & 0.05 & 0.098 & 0.00 & 0.014 & 12k \\
Beta(1,1) & 0.050 & 0.048 & 0.10 & 0.098 & 0.00 & 0.017 & 10k \\
Beta(1,1) & 0.050 & 0.050 & 0.05 & 0.100 & 0.01 & 0.014 & 11k \\
Beta(1,1) & 0.050 & 0.050 & 0.10 & 0.100 & 0.00 & 0.022 & 10k \\
Beta(1,1) & 0.100 & 0.010 & 0.05 & 0.110 & 0.02 & 0.030 & 71k \\
Beta(1,1) & 0.100 & 0.010 & 0.10 & 0.110 & 0.06 & 0.044 & 69k \\
Beta(1,1) & 0.100 & 0.050 & 0.05 & 0.150 & 0.00 & 0.037 & 10k \\
Beta(1,1) & 0.100 & 0.050 & 0.10 & 0.150 & 0.00 & 0.033 & 7k \\
Beta(1,2) & 0.025 & 0.063 & 0.05 & 0.221 & 0.00 & 0.044 & 10k \\
Beta(1,2) & 0.025 & 0.063 & 0.10 & 0.221 & 0.01 & 0.061 & 10k \\
Beta(1,2) & 0.050 & 0.010 & 0.05 & 0.234 & 0.03 & 0.075 & 79k \\
Beta(1,2) & 0.050 & 0.010 & 0.10 & 0.234 & 0.10 & 0.093 & 65k \\
Beta(1,2) & 0.050 & 0.050 & 0.05 & 0.274 & 0.01 & 0.069 & 10k \\
Beta(1,2) & 0.050 & 0.050 & 0.10 & 0.274 & 0.03 & 0.094 & 11k \\
Beta(1,2) & 0.050 & 0.091 & 0.05 & 0.314 & 0.01 & 0.077 & 5k \\
Beta(1,2) & 0.050 & 0.091 & 0.10 & 0.314 & 0.00 & 0.091 & 5k \\
Beta(1,2) & 0.100 & 0.010 & 0.05 & 0.326 & 0.04 & 0.123 & 63k \\
Beta(1,2) & 0.100 & 0.010 & 0.10 & 0.326 & 0.06 & 0.136 & 60k \\
Beta(1,2) & 0.100 & 0.050 & 0.05 & 0.366 & 0.00 & 0.113 & 10k \\
Beta(1,2) & 0.100 & 0.050 & 0.10 & 0.366 & 0.05 & 0.139 & 10k \\
Beta(1,3) & 0.025 & 0.076 & 0.05 & 0.368 & 0.00 & 0.132 & 12k \\
Beta(1,3) & 0.025 & 0.076 & 0.10 & 0.368 & 0.00 & 0.142 & 10k \\
Beta(1,3) & 0.050 & 0.010 & 0.05 & 0.378 & 0.01 & 0.176 & 87k \\
Beta(1,3) & 0.050 & 0.010 & 0.10 & 0.378 & 0.10 & 0.195 & 82k \\
Beta(1,3) & 0.050 & 0.050 & 0.05 & 0.418 & 0.00 & 0.166 & 13k \\
Beta(1,3) & 0.050 & 0.050 & 0.10 & 0.418 & 0.06 & 0.216 & 14k \\
Beta(1,3) & 0.050 & 0.096 & 0.05 & 0.464 & 0.00 & 0.183 & 7k \\
Beta(1,3) & 0.050 & 0.096 & 0.10 & 0.464 & 0.00 & 0.196 & 6k \\
Beta(1,3) & 0.100 & 0.010 & 0.05 & 0.474 & 0.06 & 0.233 & 69k \\
Beta(1,3) & 0.100 & 0.010 & 0.10 & 0.474 & 0.06 & 0.251 & 53k \\
Beta(1,3) & 0.100 & 0.050 & 0.05 & 0.514 & 0.01 & 0.220 & 10k \\
Beta(1,3) & 0.100 & 0.050 & 0.10 & 0.514 & 0.03 & 0.241 & 10k \\
\hline
\end{tabular}
\end{sc}
\end{small}
\end{center}
\vskip -0.1in
\end{table}

\end{document}